\documentclass{article}

\usepackage{microtype}
\usepackage{graphicx}
\usepackage{booktabs} 

\usepackage{hyperref}

\usepackage[accepted]{icml2023}

\usepackage{amsmath}
\usepackage{amssymb}
\usepackage{mathtools}
\usepackage{amsthm}
\usepackage{dblfloatfix}

\usepackage{graphicx}
\usepackage{caption}
\usepackage{subcaption}
\usepackage{bbm}
\usepackage{comment}
\usepackage{wrapfig}

\usepackage[capitalize,noabbrev]{cleveref}

\theoremstyle{plain}
\newtheorem{theorem}{Theorem}[section]

\theoremstyle{definition}

\theoremstyle{remark}

\usepackage[textsize=tiny]{todonotes}

\icmltitlerunning{Cooperation in the Latent Space: The Benefits of Adding Mixture Components in Variational Autoencoders}

\begin{document}

\twocolumn[
\icmltitle{Cooperation in the Latent Space: The Benefits of\\ Adding Mixture Components in Variational Autoencoders}



\icmlsetsymbol{equal}{*}

\begin{icmlauthorlist}
\icmlauthor{Oskar Kviman}{yyy,comp}
\icmlauthor{Ricky Molén}{yyy,comp}
\icmlauthor{Alexandra Hotti}{yyy,comp,klarna}
\icmlauthor{Semih Kurt}{yyy,comp}
\icmlauthor{Víctor Elvira}{xxx}
\icmlauthor{Jens Lagergren}{yyy,comp}
\end{icmlauthorlist}

\icmlaffiliation{yyy}{KTH Royal Institute of Technology}
\icmlaffiliation{comp}{Science for Life Laboratory}
\icmlaffiliation{klarna}{Klarna}
\icmlaffiliation{xxx}{University of Edinburgh}

\icmlcorrespondingauthor{Oskar Kviman}{okviman@kth.se}
\icmlcorrespondingauthor{Ricky Molén}{rickym@kth.se}

\icmlkeywords{Machine Learning, ICML}

\vskip 0.3in
]



\printAffiliationsAndNotice{} 

\begin{abstract}
In this paper, we show how the mixture components cooperate when they jointly adapt to maximize the ELBO.
We build upon recent advances in the multiple and adaptive importance sampling literature. We then model the mixture components using separate encoder networks and show empirically that the ELBO is monotonically non-decreasing as a function of the number of mixture components. These results hold for a range of different VAE architectures on the MNIST, FashionMNIST, and CIFAR-10 datasets. In this work, we also demonstrate that increasing the number of mixture components improves the latent-representation capabilities of the VAE on both image and single-cell datasets.  This cooperative behavior motivates that using Mixture VAEs should be considered a standard approach for obtaining more flexible variational approximations. Finally, Mixture VAEs are here, for the first time, compared and combined with normalizing flows, hierarchical models and/or the VampPrior in an extensive ablation study. Multiple of our Mixture VAEs achieve state-of-the-art log-likelihood results for VAE architectures on the MNIST and FashionMNIST datasets. The experiments are reproducible using our code, provided \url{https://github.com/lagergren-lab/mixturevaes}.
\end{abstract}

\section{Introduction}
\label{sec:intro}
Adaptive importance sampling (AIS) methods aim at approximating a target distribution by a set of proposal densities, often interpreted as a mixture density \citep{elvira2019generalized}. However, choosing appropriate proposals is known to be a difficult problem, so they are sequentially updated according to an adaptation scheme to reduce the mismatch with the target. The adaptive importance sampling methodology \citep{bugallo2017adaptive} is currently enjoying a great amount of research interest, e.g. regarding its convergence guarantees \citep{akyildiz2021convergence, akyildiz2022global}, and is considered a promising contender for the MCMC methodology \citep{owen2013monte,luengo2020survey}.

Meanwhile, in variational inference (VI), the mismatch between a variational approximation and the target distribution is minimized via optimization of, typically, the Kullback-Leibler (KL) divergence. If one uses this optimization scheme to adapt the proposal densities in AIS, VI appears to be a special case in the AIS methodology. However, whereas VI is well-known in the machine learning (ML) community, AIS has received little attention until very recently.

The signal processing literature has recently started to see an increase of studies at the intersection between AIS and VI \citep{wang2019adaptive, wang2021adaptive}. For example, in \citet{el2019variational}, they obtain a mixture of proposal distributions via auto-differentiation VI (ADVI; \citet{kucukelbir2017automatic} as a starting point for an AIS algorithm. After the adaptation, the mixture of proposals is again updated via ADVI. 

Similar to AIS, mixture distributions in VI and variational autoencoders (VAEs; \citet{kingma2013auto}) is a research area which is also receiving a lot of attention  \citep{nalisnick2016approximate, kucukelbir2017automatic, morningstar2021automatic, kviman2022multiple} since they result in increased performance due to more flexible posterior approximations. Despite this, mixture models in VAEs are not considered standard tools. Instead, one would typically use normalizing flows (NFs; \citep{rezende2015variational, papamakarios2021normalizing}), hierarchical models \citep{burda2015importance, sonderby2016ladder,vahdat2020nvae}, autoregressive models \citep{van2016pixel} or more complex prior distributions \citep{tomczak2018vae, bauer2019resampled, aneja2021contrastive}. In this work, we argue that mixture models are indeed off-the-shelf solutions in VAEs. To that effect, we demonstrate that mixture models result in increased flexibility when combined with all of the four standard methods mentioned above. We also share our in-detail implementation instructions on how to combine mixtures with other popular tools in what we refer to as the \textit{mixture cookbook}, found in Sec. \ref{sec:cookbook}. By doing so, we attempt to lower the barriers to using mixture models in VAEs.

The concepts of the Multiple Importance Sampling Evidence Lower Bound (MISELBO) and ensembles of variational approximations were established in \citet{kviman2022multiple}.  MISELBO offers a simple way to compute the ELBO for mixtures or ensembles inspired by the MIS literature (more in Sec. \ref{sec:miselbo}). 
 To disambiguate between mixtures and ensembles, we refer to ensembles as compositions of independently learned approximations of the same target density (as in \citet{kviman2022multiple}), while mixtures will be considered to have learned their components jointly. When applied to VAEs, we call them \textit{Ensemble VAEs} and \textit{Mixture VAEs}, respectively. Moreover, we show in Sec. \ref{sec:2d experiments} and Sec. \ref{sec:latent images} that ensemble learning can lead to loss of diversity, while learning with MISELBO enforces diversity, and connect this to findings in the AIS literature.
 
 Using our mixture cookbook, we propose a  \textit{composite model}; a hierarchical Mixture VAE with NFs, a VampPrior \citep{tomczak2018vae} and a PixelCNN decoder \citep{van2016pixel}. In Sec. \ref{sec:components} we incrementally construct this model and successively evaluating the effect of adding each of the encoder-related architectures for different number of mixture components, $S$. The pattern is clear: adding mixture components improves the marginal log-likelihood estimates for all the considered models. Also, multiple of the Mixture VAEs, including our composite model, achieve state-of-the-art scores on the MNIST dataset when compared to existing VAE architectures.

\begin{figure*}
\centering
\begin{subfigure}{0.19\textwidth}

    \includegraphics[width=\textwidth]{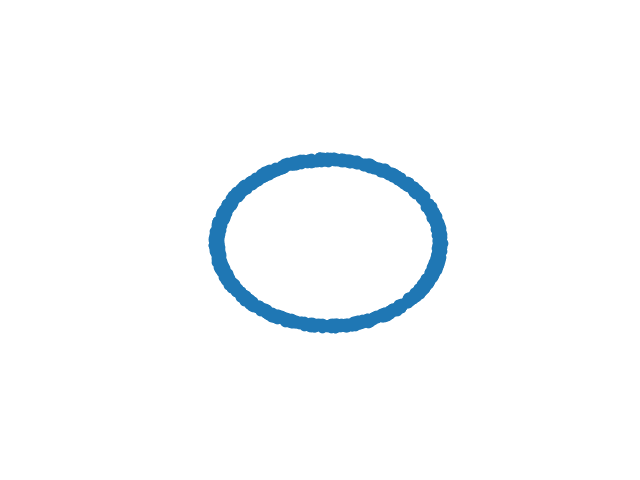}
    \label{fig:twoD_a1}
\end{subfigure}%
~
\begin{subfigure}{0.19\textwidth}

    \includegraphics[width=\textwidth]{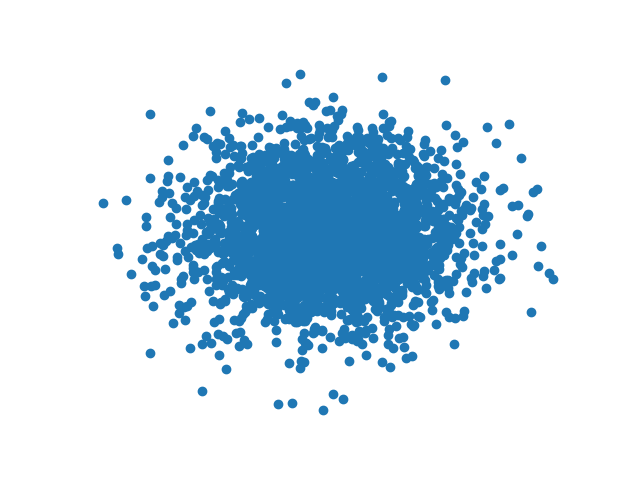}
    \label{fig:twoD_b1}
\end{subfigure}%
~
\begin{subfigure}{0.19\textwidth}

    \includegraphics[width=\textwidth]{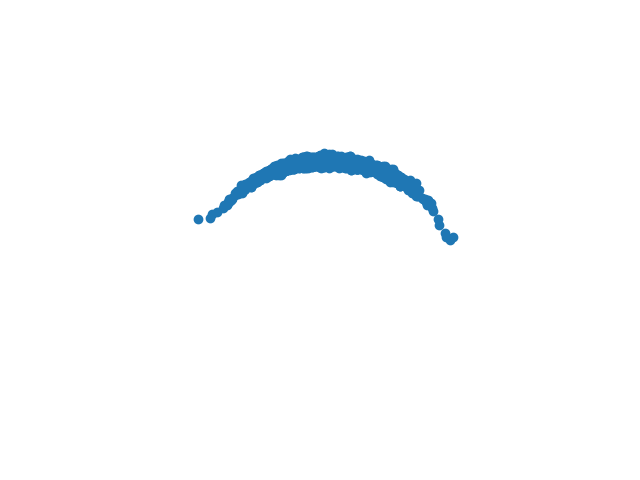}
    \label{fig:twoD_c1}
\end{subfigure}%
~
\begin{subfigure}{0.19\textwidth}
    \includegraphics[width=\textwidth]{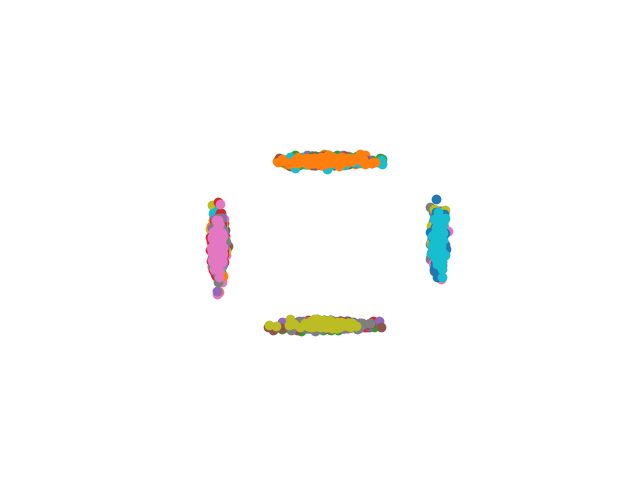}
    \label{fig:twoD_d1}
\end{subfigure}%
~
\begin{subfigure}{0.19\textwidth}
    \includegraphics[width=\textwidth]{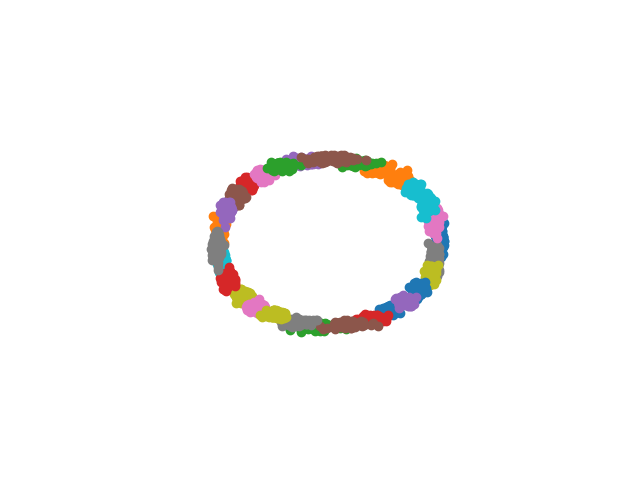}
    \label{fig:twoD_e1}
\end{subfigure}

\begin{subfigure}{0.19\textwidth}
    \includegraphics[width=\textwidth]{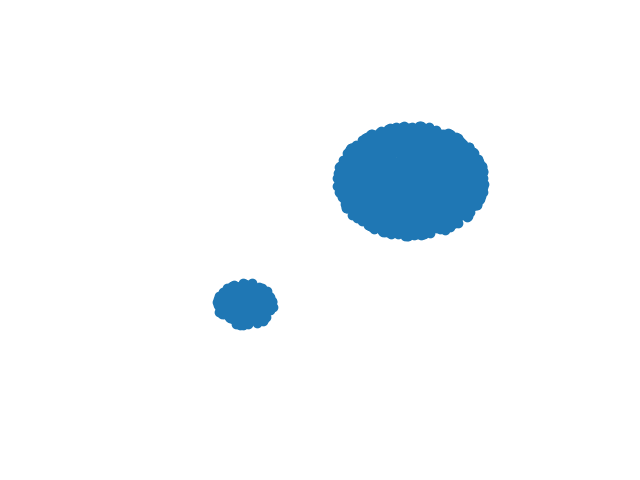}
    \label{fig:twoD_a}
\end{subfigure}%
~
\begin{subfigure}{0.19\textwidth}
    \includegraphics[width=\textwidth]{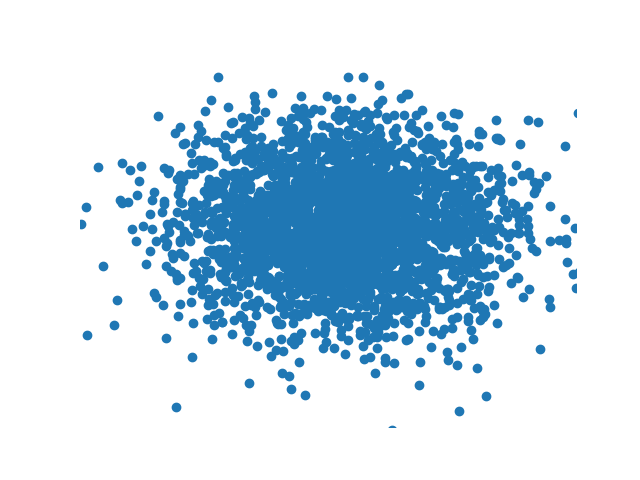}
    \label{fig:twoD_b}
\end{subfigure}%
~
\begin{subfigure}{0.19\textwidth}
    \includegraphics[width=\textwidth]{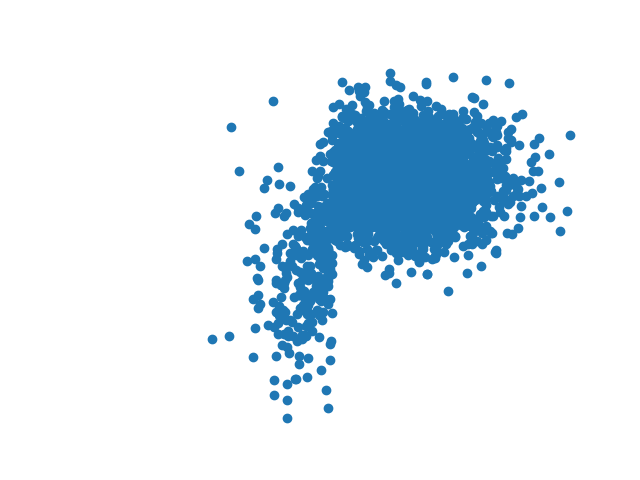}
    \label{fig:twoD_c}
\end{subfigure}%
~
\begin{subfigure}{0.19\textwidth}
    \includegraphics[width=\textwidth]{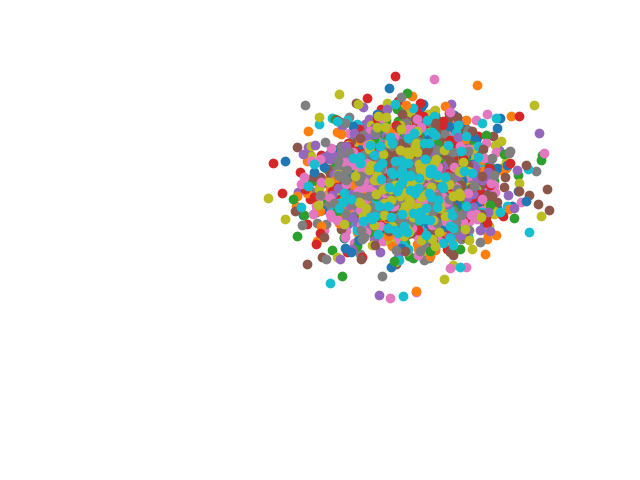}
    \label{fig:twoD_d}
\end{subfigure}%
~
\begin{subfigure}{0.19\textwidth}
    \includegraphics[width=\textwidth]{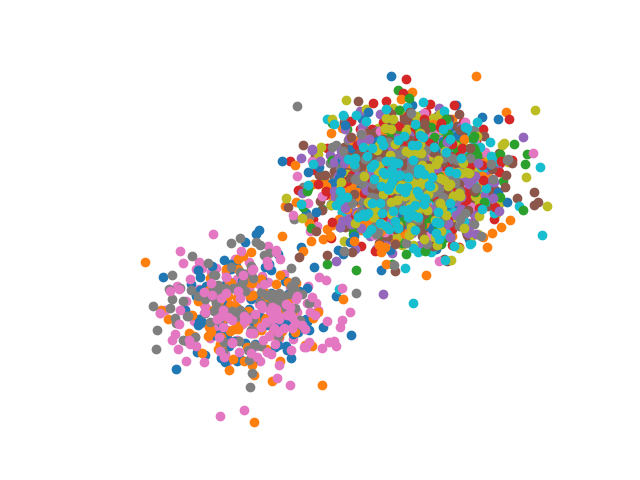}
    \label{fig:twoD_e}
\end{subfigure}
        
\caption{From left to right, applying to both rows: The true target distribution is shown in the leftmost column, followed by a single approximation learned with an importance weighted version of the KL divergence, an IAF approximation, an ensemble approximation and, finally, a mixture approximation. The mixture is the only approximation capable of covering the target distribution. Colors represent samples from different components.}
\label{fig:twoD}
\end{figure*}
 
 Finally, we show that the latent representations learned with Mixture VAEs are superior to those learned with Ensemble VAEs or the Vanilla VAE, measured on the downstream task of linear classification. Apart from images, the classification is also performed on single-cell transcriptome data by applying Ensemble and Mixture VAEs to the single-cell VI algorithm  \citep{lopez2018deep}.
 
 We now summarize our contributions as follows:
 \begin{itemize}
     \item In Sec.  \ref{sec:monotone}, we empirically show that the ELBO scores are monotonically non-decreasing as we add more mixture components.
     \item We make Mixture VAEs more accessible with the mixture cookbook, which shows how to use a mixture of encoders in conjunction with three popular VAE architectures, namely hierarchical models, NFs, and the VampPrior (Sec. \ref{sec:cookbook}).
     \item We show that learning with MISELBO is strongly connected to the adaptive mechanism that AIS algorithms use in order to learn the proposals.
     \item Based on the connection between AIS and VI, we argue (Sec. \ref{sec:ais+miselbo}) and show (Sec. \ref{sec:2d experiments}) that the mixture components in a Mixture VAE complement each other by jointly distributing themselves to cover the entire target distribution. 
    
     \item We put forth a novel composite model: a Hierarchical Mixture VAE with NFs, a VampPrior and a PixelCNN decoder. The composite model outperforms the majority of existing VAE architectures in terms of log-likelihood estimates on the MNIST dataset.
 \end{itemize}

 \section{Background}
\label{sec:background}
The main topics in this work are AIS, MISELBO and VAEs/amortized VI. In this section, we cover the first two as we expect that these are the least familiar concepts to the reader. For thorough descriptions of VAEs or amortized VI, we refer to \citet{kingma2019introduction} or \citet{zhang2018advances}, respectively. We end the section by drawing connections between AIS and MISELBO.

\subsection{Adaptive Importance Sampling}
\label{sec:ais}
AIS is a principled methodology for learning proposal densities in importance sampling (IS). The goal is generally learning the parameters of such proposals. There exist several families of AIS algorithms, depending on how the parameters are learnt. A common approach is to rely on the simulated samples and their IS weights in order to perform moment matching estimators \citep{cornuet2012,el2019efficient} or via resampling schemes \citep{cappe04,cappe08,elvira2022optimized}. Let $z$ be a latent variable and $x$ an observation, then the IS weights are given by
\begin{equation}
\label{eq:is weigths}
    w^{L}_\text{IS} = \frac{1}{L}\sum_{\ell=1}^L \frac{p_\theta(x,z_{\ell})}{q_{\phi}(z_{\ell}|x)}\quad z_{\ell}\sim q_{\phi}(z|x),
\end{equation}
if only one proposal $ q_{\phi}(\cdot|x)$ is available, or by
\begin{equation}
\label{eq:mis weigths}
    w^{L,s}_\text{MIS} = \frac{1}{L}\sum_{\ell=1}^L \frac{p_\theta(x,z_{s,\ell})}{\sum_{j=1}^S \pi_j q_{\phi_j}(z_{s,\ell}|x)} \quad z_{s,\ell}\sim q_{\phi_s}(z|x)
\end{equation}
if a mixture of proposals is used to simulate the samples (see alternative weighting schemes in \citet{elvira2019generalized}). Here, $q_\phi(z|x)$ can be any distribution that is appropriate w.r.t. target distribution $p_\theta(z|x)$.

%
The generality of the adaptation scheme provides the practitioner with high modeling flexibility. For instance, common adaptation schemes are gradient based \citep{elvira2015gradient,schuster15} or resampling based \citep{cappe04,cappe08} (or both \citep{elvira2022optimized}). As a particular example, in deterministic mixture population Monte Carlo (DM-PMC;  \citet{elvira2017improving}), an equally weighted mixture proposal is iteratively adapted via sampling, weighting, and resampling steps. Unlike previous schemes, the importance weights are constructed by using the whole mixture in the denominator, i.e., as in the denominator of Eq.~\ref{eq:mis weigths}. This weighting scheme has shown superior performance both for reducing the variance of the estimators \citep{elvira2019generalized} and for an increased diversity in the adaptive mechanism \citep{elvira2017population}. 

\subsection{MISELBO}
\label{sec:miselbo}
For a single mixture component, a tighter bound on the marginal log-likelihood than the vanilla ELBO, $\mathcal{L}_\textnormal{ELBO}$, is given by the importance weighted ELBO (IWELBO; \citet{burda2015importance}) as
\begin{equation}
\label{eq:iwelbo}
    \mathcal{L}^L_\textnormal{IWELBO}(x; q_{\phi}) =  \mathbb{E}_{q_{\phi}(z|x)}\left[
    \log \frac{1}{L}\sum_{\ell=1}^L \frac{p_\theta(x, z_{\ell})}{q_{\phi}(z_{\ell}|x)}
    \right],
\end{equation}
where $\mathcal{L}_\textnormal{ELBO}$ is retrieved by letting $L=1$. 
%
%
Recently \citet{kviman2022multiple} drew a connection between the extension of Eq. \eqref{eq:iwelbo} to mixture approximations, and MIS. This resulted in an MIS scheme for computing the ELBO (where simulations are drawn from all mixture components without replacement), leading to the MISELBO objective:\footnote{An alternative name of the same lower bound arises naturally in the work of \citet{morningstar2021automatic} where the sampling scheme is instead viewed as stratified sampling, resulting in the name SELBO or SIWELBO.}
\begin{align}
\label{eq:miselbo}
    &\mathcal{L}^L_\textnormal{MIS}(x; q_S(z|x)) =\\&\frac{1}{S}\sum_{s=1}^S \mathbb{E}_{q_{\phi_s}(z|x)}\left[
    \log \frac{1}{L}\sum_{\ell=1}^L \frac{p_\theta(x, z_{s,\ell})}{\frac{1}{S}\sum_{j=1}^S q_{\phi_j}(z_{s,\ell}|x)}
    \right].\nonumber
\end{align}
When $S$ components are learned independently, it has been shown that forming a uniform mixture and joining their contributions via MISELBO offers a tighter bound than when averaging their individual vanilla-ELBO contributions \citep{kviman2022multiple}. 

\subsection{An Adaptive Importance Sampling View of MISELBO}
\label{sec:ais+miselbo}
A connection between MIS and MISELBO has been recently established in \citet{kviman2022multiple}.
This can be seen, for instance by noting that  $\mathcal{L}^L_\textnormal{MIS}$ is a function of the mixture weights $w^L_\textnormal{MIS}$.
%
%
However, the vision in \citet{kviman2022multiple} is limited since the set of proposals are given by independent learning processes. 
In this paper, we take a step further and use MISELBO to also learn the mixture components. We show that this approach is strongly connected to the adaptive mechanism that AIS algorithms use in order to learn the proposals. 



DM-PMC mentioned in Sec. \ref{sec:ais} is an AIS method that, at each iteration, builds an unweighted mixture proposal which approximates the target distribution. Thus, there is a close connection between learning mixtures in DM-PMC (and more generally in AIS) and doing so with MISELBO, when the mixture in Eq. \eqref{eq:miselbo} is equally weighted. The resemblance lies in how the samples are simulated from the mixture, but also in that the parameters of the proposals/variational approximations are adapted based on the mixture weights $w_\text{MIS}^L$. The difference is that, in DM-PMC, the parameters are adapted via resampling, whereas VI is gradient based. 
%

\begin{figure}
    \centering
\includegraphics[width=0.4\textwidth]{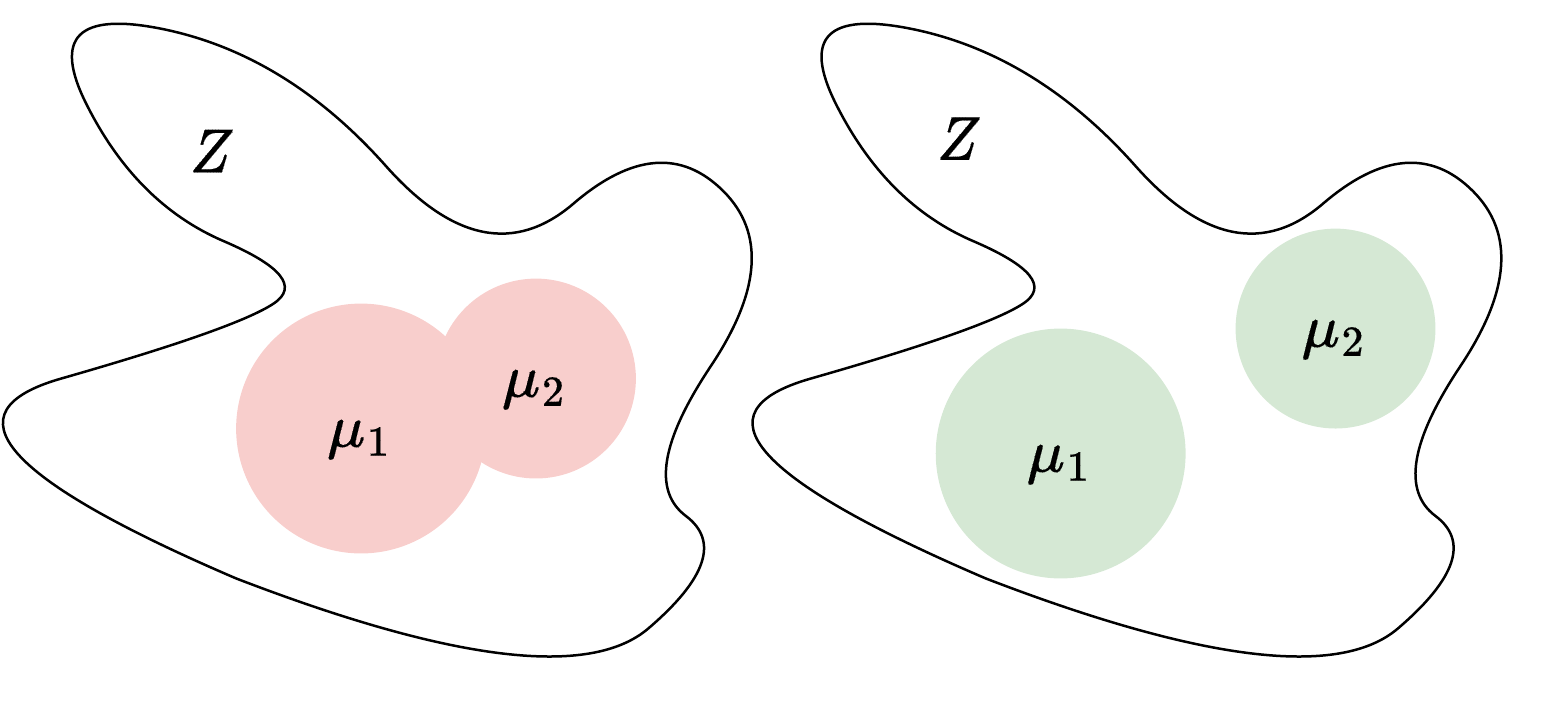}
    \caption{Consider two different mixture approximations of $p_\theta(z|x)$ in $\mathcal{Z}$, the red one in the left plot and the green one to the right. The radius of the disks is of length one $\sigma$. If the numerator in Eq. \eqref{eq:miselbo} is unchanged when comparing the two approximations, the green mixture will obtain a higher ELBO. This phenomenon is explained in Sec. \ref{sec:ais+miselbo}.}
\label{fig:green_mixture}
\end{figure}

We now exploit the connection with AIS to better understand the behavior of the mixture components learned via MISELBO. In \citet{elvira2017population}, it was shown that due to the use of the mixture in the denominator of the importance weight,  $w_\text{MIS}^L$, the components cooperate in order to cover the target distribution. Studying the denominator in Eq. \eqref{eq:miselbo}, this insight is clearly applicable in the VI setting; the more separated the components are, the smaller the denominator gets. Furthermore, the insight inspired our experiment setup in Sec. \ref{sec:2d experiments}, and the resulting cooperative behaviour is depicted in Fig. \ref{fig:green_mixture}.
%

In contrast, when adapting $S$ proposal densities using $w_\text{IS}^L$, the components can not cooperate and thus result in a worse approximations of the target  \cite{elvira2017improving,elvira2017population}. In VI, this corresponds to learning ensembles of variational approximations \cite{kviman2022multiple}.

Based on these findings from the AIS literature, we expect that the same will happen in our experiments here. That is for a mixture learned with MISELBO, the components will cooperate to cover the target, while an ensemble learned with IWELBO or the ELBO will not. We will see that this is indeed the case in Sec. \ref{sec:2d experiments}.


As in \citet{cappe04,elvira2017population,elvira2022optimized}, we focus on uniform mixtures, which hold the benefit of not needing to select the mixture weights, simplifying also the sampling process. Uniform mixture weights are also motivated from the perspective of general applicability to VAEs. With uniform weights, the architecture of the encoder network does not need to be revised in order to map $x$ to the mixture weights (e.g., as in \citet{nalisnick2016approximate}).

Generally, drawing connections between separate research fields is an effective way to obtain inspiration to advance existing methods. By highlighting the connection between VI and AIS, we open up the possibility of exploiting the vast AIS literature \citep{bugallo2017adaptive}. Examples of interesting directions to explore, outside the scope of this work, are both theoretical (such as utilizing established convergence guarantees in AIS for parametric proposal distributions \citep{akyildiz2021convergence}) and empirical (e.g., exploring different weighting schemes of the mixture distributions \citep{bugallo2017adaptive}).

\section{The Mixture Cookbook}
\label{sec:cookbook}
%
Here we propose the first systematic approach to optimize mixtures in three of the most popular VAE models, namely in (1) hierarchical models, (2) normalizing flows, and (3) VampPrior models. 
We dedicate a subsection to each of the tools, giving the explicit mixture-related notation that will help practitioners to avoid non-trivial pitfalls. In the following subsections, we consider the case with $L=1$ without loss of generality, in order to avoid cluttered notation and since this is the most common setting during training.

\begin{table*}[!b]
  \caption{The NLL results for an increasing number of mixtures ($S$) for the Mixture VAEs considered in Sec. \ref{sec:components} on the MNIST dataset. Our composite model (bottom row) is a composition of the other four models. A lower NLL metric is better, and the results are averages using three random seeds.}
  \label{tab:NLL big table}
  \centering
  \begin{tabular}{llllll}
    \toprule

    Model     & $S=1$     & $S=2$ & $S=3$     & $S=4$   \\
    \midrule
    Vanilla VAE & 79.74 $\pm$ 0.03  & 79.15 $\pm$ 0.03  & 78.87 $\pm$ 0.03 & 78.60 $\pm$ 0.05 \\
    Hierarchical VAE     & 78.92 $\pm$ 0.08 & 78.25 $\pm$ 0.04 & 77.95 $\pm$ 0.04 & 77.75 $\pm$ 0.03  \\
    VAE w. NF     &  79.43 $\pm$ 0.03  & 78.85 $\pm$ 0.05  & 78.47   $\pm$ 0.02 & 78.23 $\pm$ 0.03 \\
    VAE w. VampPrior     & 78.69 $\pm$ 0.02     &  78.06 $\pm$ 0.02 & 77.74 $\pm$ 0.02 & 77.55 $\pm$ 0.05\\
    Composite Model & 78.45 $\pm$ 0.11  &  77.80 $\pm$ 0.14    &  77.55 $\pm$ 0.11 & \textbf{77.23} $\pm$ \textbf{0.11} \\
    \bottomrule
  \end{tabular}
\end{table*}
\subsection{How to Apply Hierarchical Models}
Let  $H$ be the number of levels, then the sample from the $h$th layer is $z_s^h \sim q_{\phi_s^h}(z^h|z^{h +1})$. We compute the likelihood of the samples $z_s^1, ..., z_s^H$ with respect to the $j$th hierarchical variational approximation  according to 
\begin{equation}
    q_{\phi_j}(\mathbf{z_s}|x) = q_{\phi^H_j}(z^H_s|x)\prod_{h=1}^{H-1} q_{\phi^h_j}(z^{h}_s|z^{h+1}_s).
\end{equation}
Then, the hierarchical version of Eq. \eqref{eq:miselbo} is
\begin{align}
    \label{eq:hierarchy miselbo}
    &\mathcal{L}_\textnormal{MIS} =\\ &\frac{1}{S}\sum_{s=1}^S \mathbb{E}_{z_{s}^1,...,z_{s}^H \sim q_{\phi_s}(\mathbf{z_s}|x)}\left[
    \log \frac{p_\theta(x, \mathbf{z_{s}})}{\frac{1}{S}\sum_{j=1}^S q_{\phi_j}(\mathbf{z_{s}}|x)}
    \right].\nonumber
\end{align}
\textbf{Key observation.} In the $s$th term of Eq. \eqref{eq:hierarchy miselbo}, for each layer $h$, the samples must be drawn from the $s$th variational approximation. As a rule of thumb, all samples that appear in the $s$th term need to come from encoder $s$ for the expectation to be properly evaluated.

\subsection{How to Apply Normalizing Flows}
In normalizing flows, a sample is drawn from a base distribution and then passed through the flow, a sequence of $T$ deterministic and bijective transformations, $\{f_{\gamma_s^t}\}_{t=1}^T$, where $\gamma_s^t$ are the learnable parameters of the $t$th mapping specific for the $s$th encoder network. We here let $q_{\phi_{s}}(z|x)$ denote the base distribution and $z^T_{s}$ be the output of the $T$ long flow with $z_s:= z_{s}^0$ as input. However, as explained in the paragraph below, evaluating the likelihood of $z^T_{s}$ on another NF-based mixture component is complicated, and costly in terms of run time and memory requirements. To solve this issue, one can instead use a single flow of transformations, $\{f_{\gamma^t}\}_{t=1}^T$. If the transformations are chosen such that the corresponding Jacobian matrix is upper or lower triangular\footnote{We direct the reader to \citet{papamakarios2021normalizing} for in-depth details of NFs.}, we can, via the change-of-variables formula, express the $j$th mixture member's resulting density's goodness of fit to the sample from component $s$ as
\begin{equation}
\label{eq:nf variational approximation}
    q_{\phi^T_j}(z^{T}_{s}|x) = q_{\phi_j}(z^0_s|x)\prod_{t=1}^T\left\vert \frac{df_{\gamma^t}(z_{s}^{t-1})}{dz_{s}^{t-1}}\right\vert^{-1}.
\end{equation}

\textbf{Key observation.} Evaluating the likelihood of $z^T_{s}$ on another NF-based mixture component is complicated in practice, as it would require $T$ inverse transformations of $z^T_{s}$. Instead, if the flow model is shared, the sequence of transformed latent variables is known, and so the product in Eq. \eqref{eq:nf variational approximation} only needs to be computed once (see Eq. \eqref{eq:miselbo_composite} for an example of the resulting MISELBO expression when the flow is shared). This, in turn, drastically reduces memory requirements and run time, while sacrificing flexibility of the approximations. Still, the base distributions will be encouraged to collaborate in order to minimize the denominator in Eq. \eqref{eq:miselbo}.


\subsection{How to Apply the VampPrior}
The VampPrior models the prior as the variational approximation aggregated over $K$ \textit{pseudo inputs}, $u_k$, learned representations of the data. If the parametric form of the variational approximation is Gaussian, then the VampPrior is a mixture of Gaussians conditioned on different (pseudo) inputs.

We now show how to apply the VampPrior to ensembles of variational approximations.  We get
\begin{equation}
\label{eq:vampprior}
    p_\lambda(z) = \frac{1}{K} \sum_{k=1}^K q_\phi(z|u_{k}) = \frac{1}{KS} \sum_{k=1}^K\sum_{s=1}^S q_{\phi_s}(z|u_{k}),
\end{equation}
where the first equality is the definition of the VampPrior for a generic variational approximation, whereas the second comes from using a mixture approximation. The pseudo inputs are generated by a neural network with learnable parameters and takes a $K$-dimensional identity matrix as input. 

\textbf{Key observation.} We cannot use $S$ separate prior distributions, one for each variational approximation. This is crucial, as it would violate the criterion that all variational approximations need to share the same target distribution \citep{kviman2022multiple}. In practice, one way to achieve this is by using a single pseudo-input generator and computing the above expression.

\begin{table*}
  \caption{The NLL and bits-per-dimension results for an increasing number of mixtures ($S$) for a subset of the Mixture VAEs considered in Sec. \ref{sec:components} on the FashionMNIST and CIFAR-10 datasets. The lower the better.}
  \label{tab:NLL big table FashionMNIST}
  \centering
  \begin{tabular}{lllllll}
    \toprule

    Dataset & Model     & $S=1$     & $S=2$ & $S=3$     & $S=4$   \\
    \midrule
    FashionMNIST & Vanilla VAE & 225.50 & 224.715 & 224.39  & 224.22  \\
    FashionMNIST & Composite Model & 223.58   &  223.07    &  222.62 &\textbf{222.38}  \\
    \midrule
    CIFAR-10 & Vanilla VAE & 4.85 & 4.85 & 4.84  & \textbf{4.83}  \\
    \bottomrule
  \end{tabular}
\end{table*}

\begin{table*}[!b]
  \caption{NLL statistics for different state-of-the-art VAE architectures on the MNIST dataset.}
  \label{tab:NLL SOTA}
  \centering
  \begin{tabular}{ll}
    \toprule

    Model     & NLL   \\
    \midrule
    Hierarchical VAE w. VampPrior \citep{tomczak2018vae} & 78.45\\
    NVAE \citep{vahdat2020nvae} & 78.01 \\
    PixelVAE++ \citep{sadeghi2019pixelvae++} & 78.00 \\
    MAE \citep{ma2019mae} & 77.98 \\
    Ensemble NVAE \citep{kviman2022multiple} & 77.77 $\pm$ 0.2\\
    Vanilla VAE ($S=12$; \textbf{our}) & 77.67 \\
    Composite Model ($S=4$; \textbf{our})     &  \textbf{77.23 $\pm$ 0.1}  \\
    \bottomrule
  \end{tabular}
\end{table*}

\section{Experiments}

\subsection{Two-Dimensional Examples}
\label{sec:2d experiments}
We now demonstrate the benefits of the type of flexibility obtained with mixtures for density estimation, contrasted to an importance weighted approximation (IWVI), IAF, and ensembles.
We show that, in contrast to standard ensemble strategies, the coordinated learning of the mixture allows for the cooperation of the components to jointly approximate the target distribution.
This can be easily understood by highlighting the connection between our strategy and relevant AIS algorithms, as discussed in Sec.\ref{sec:ais+miselbo}. 
%
%
The expected cooperative behavior holds in two difficult cases: (i) when $p(z)$ is a ring-shaped density, and (ii) when the target is an unevenly weighted bi-modal Gaussian, as in the second row in Fig. \ref{fig:twoD}.(a) where we have   $p(z)=0.8\mathcal{N}(\mathbf{1}, 0.1I_2) + 0.2\mathcal{N}(-\mathbf{1}, 0.1I_2)$. 

In Fig. \ref{fig:twoD} we compare an $S=30$-component mixture (e) to ensembles of variational approximations (d) with $S=30$, trained with $L=1$; an inverse autoregressive flow (IAF; \citet{kingma2016improved}) (c) with $T=30$ flows and a two-layered MADE \citep{germain2015made} with 10 hidden units in each layer; IWVI (b), a single variational approximation with $L=30$; the true distributions, $p(z)$. The IAF is by far the most advanced model of the four approximation methods, and was given time to converge. All approximations and the base distribution for IAF were Gaussian or a mixture of Gaussian's with learnable parameters. See Sec. \ref{sec:app 2d experiments} for more details.

The results demonstrate that the mixture coordinates its components to spread out and cover $p(z)$, while the ensemble overpopulates certain parts of the target. For example in the bimodal Gaussian case (lower row in Fig. \ref{fig:twoD}), the mixture distributes its components between the modes proportionally to  $p(z)$. Meanwhile, the ensemble fails to cover the lighter mode (lower row, second to right). Interestingly, using $L=30$ for a single approximation seems to cause it to become indecisive (Fig. \ref{fig:twoD}, second column to the left). This does not happen when $L=1$, as is clear in the ensemble case. Finally, despite the large complexity of the IAF, it was not flexible enough to capture the target distributions (Fig. \ref{fig:twoD}, middle column).

\subsection{Log-Likelihood Estimation}
Here we use the marginal log-likelihood in order to quantify the effect of applying Mixture VAEs to a set of popular VAE algorithms. This is easily achieved through the guidance of our mixture cookbook in Sec. \ref{sec:cookbook}. Inspired by \citet{kviman2022multiple}, the mixture components were modelled with separate encoder nets. In the final subsection, we critically analyze the source of the results in Sec. \ref{sec:components} with some adversary examples. See Sec. \ref{sec:app ll} for in-depth experimental details.

\subsubsection{Ablation Study}
\label{sec:components}
In this section, we incrementally construct \textit{the composite model}, i.e. a Mixture VAE, which incorporates all popular techniques discussed in this work; hierarchical models, NFs and the VampPrior. Consequently, we perform a large ablation study consisting of 20 different models. All models employ $S$ separate encoder networks, and the models that use NFs utilize shared flow models, as explained in Sec. \ref{sec:cookbook}. We ran each model three times and averaged their results. In Table \ref{tab:NLL big table}, we report the effect of applying mixtures to each of the architectures for MNIST.
\begin{table*}[!b]
  \caption{\textbf{MNIST:} Accuracy for linear classification using the latent representations for an increasing number of mixture components ($S$).}
  \label{tab: mnist}
  \centering
  \begin{tabular}{llllll}
    \toprule
    Model     & $S=1$     & $S=2$ & $S=3$     & $S=4$ & $S=5$  \\
    \midrule

    VAE  &  94.9$\pm$4E-3\% & 94.8$\pm$4E-3\% & 94.9$\pm$4E-3\% & 94.8$\pm$5E-3\% & 94.8$\pm$5E-3\%\\
    
    Ensemble VAE   & \textbf{95.5$\pm$4E-3}\% &  95.7$\pm$4E-3\% & 95.4$\pm$2E-3\% & 95.9$\pm$2E-3\% & 96.0$\pm$2E-3\% \\

    Mixture VAE & \textbf{95.5$\pm$4E-3}\% & \textbf{95.8$\pm$3E-3}\% & \textbf{96.1$\pm$2E-3}\% & \textbf{96.5$\pm$3E-3}\% & \textbf{96.6$\pm$2E-3}\%\\
    \bottomrule
  \end{tabular}
\end{table*}
The Vanilla VAE exactly followed the architecture of the PixelCNN \citep{van2016pixel} decoder + CNN encoder nets with gating mechanisms \citep{dauphin2017language} described in \citet{tomczak2018vae}, but without hierarchies. The latent space was 40 dimensions. Note that this was the base architecture for all subsequent models. The Hierarchical VAE is the two-layered PixelCNN VAE from \citep{tomczak2018vae}, without the VampPrior. In the VAE with NF, we
employed the IAF with $T=2$ flows and a two-layered MADE with 320 hidden units in each layer, as in the original paper \citep{kingma2016improved}. The VAE with VampPrior is the Vanilla VAE with the addition of a $K=500$ VampPrior \citep{tomczak2018vae}. Finally, we implement the composite model. The IAFs were placed in the
lower-level latent space, where the prior was a normal distribution with learnable parameters. The VampPrior was implemented in the upper-level latent space. The explicit form of the objective function is in Eq. \eqref{eq:miselbo_composite}. 

\subsubsection{Empirical Monotonicity of Mixture VAEs}
\label{sec:monotone}
Here we describe a novel and impactful finding. In all our experiments, we found that the NLL is a non-increasing function of $S$. Thus, to improve the NLL scores, it is sensible to simply incorporate a mixture of encoders into an existing model. The power of combining mixtures with popular VAE architectures is clearly demonstrated in tables \ref{tab:NLL big table} and \ref{tab:NLL big table FashionMNIST}. These results promote future research on Mixture VAEs, and mixture approximations in VI in general.

\begin{figure}
    \centering
\includegraphics[width=0.4\textwidth]{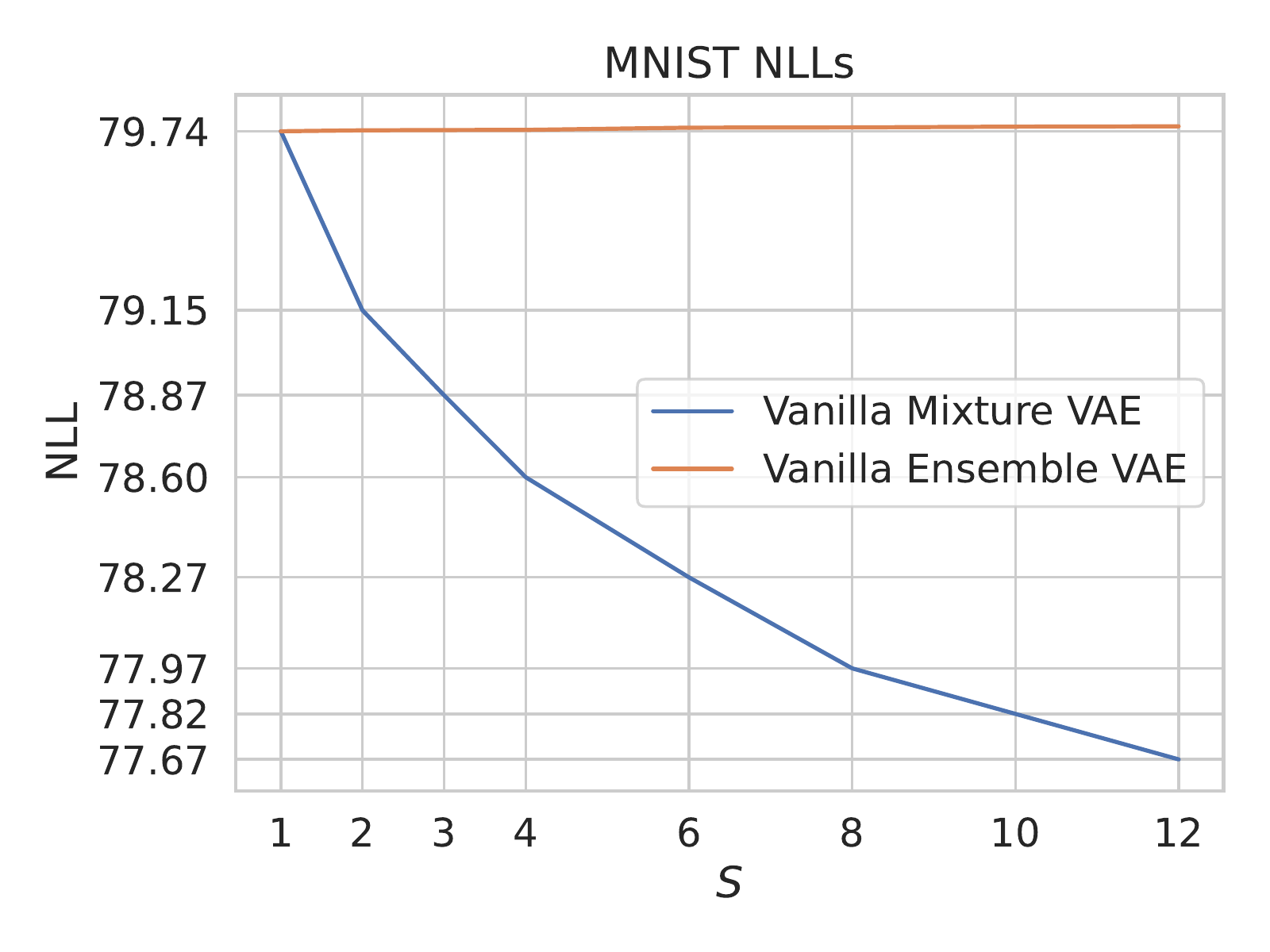}
    \caption{Monotonic NLL improvement (the lower the better) as $S$ increases. The results are produced on the MNIST dataset using the Vanilla VAE from Sec. \ref{sec:components}, and the corresponding Ensemble VAE. In this experiment, the Ensemble VAE becomes slightly worse with $S$, indicating that the ensemble components do not sufficiently diversify as a result of them being trained independently.}
    \label{fig:mnist nll monotonic}
\end{figure}

To stress test the monotonicity claim, we trained Vanilla VAEs with all the way up to $S=12$ encoder networks, see Fig. \ref{fig:mnist nll monotonic} for an illustration. For every added component, we saw an improvement in NLL. Although the relative improvement of adding more components decreased after $S=6$, and the possible improvement is upper bounded in theory, we could not find a case where larger $S$ did not give an improvement in this setup. 

Moreover, to the best of our knowledge, impressively many of the Mixture VAEs, especially our composite model, achieve state-of-the-art results on the MNIST dataset compared to other VAE architectures, as shown in Table \ref{tab:NLL SOTA}.\footnote{Explicitly comparing with VAE architectures excludes some brilliant works, such as consistency regularization in VAEs \citep{sinha2021consistency}, which is a regularizing term in the training objective function.
}


\subsubsection{Big IWAE versus Mixture VAE}
It is necessary to explore whether the improved NLL scores obtained by the Mixture VAEs for increasing S are due to the mixture approximations, rather than the increased number of parameters or the number of samples. Indeed, the number of encoder network parameters increases as a function of $S$, and the Mixture VAE uses $S\times L$ samples to approximate its objective function. The results from this comparison can be found in Table  \ref{tab:big IWAE vs MISVAE}. In general, we found that the IWAE could not outperform the corresponding Mixture VAE regardless if we increased the IWAE's number of hidden layers or its number of hidden units. Conversely, we found that naively increasing the number of parameters in the IWAE sometimes resulted in worse NLL scores. Meanwhile, naively adding mixture components works well in this regard. See Appendix \ref{sec:big_mix_appendix} for more experimental details.

\begin{table}
  \caption{Comparison between IWAEs trained with $L=S$ importance samples and at least as many network parameters as the Mixture VAEs for $S=1,2,3$. We either trained IWAE with an increasing number of hidden layers $N$ for a fixed number of hidden units $n$, or the other way around. In all cases, we round off upwards, such that the IWAEs enjoy more parameters than Mixture VAE. The performances are measured in NLL, the lower the better.}
  \label{tab:big IWAE vs MISVAE}
  \centering
  \begin{tabular}{llll}
    \toprule
    Model     & $S=1$     & $S=2$ & $S=3$\\
    \toprule
    Mixture VAE &&&\\
    \midrule
    NLL&85.24&85.18&84.77\\
    \text{\# enc. params.} & 349,200&  698,400& 1,047,600\\
    \text{\# total params.}& 790,384& 1,139,584& 1,488,784\\
    \toprule
    IWAE ($N$) &  $N=1$ & $N= 5$ & $N=9$ \\ 
    \midrule
    NLL &85.24 &  97.45 & 101.62\\
    \text{\# enc. params.} &   349,200& 709,200 & 1,069,200\\
    \text{\# total params.}& 790,384 &  1,150,384 &  1,510,384 \\
    \toprule
     IWAE ($n$) &  $n=300$ & $n=509$  & $n=679 $ \\
     \midrule
     NLL&85.24&86.28& 85.16\\
     \text{\# enc. params.}& 349,200&698,857&1,047,697\\
     \text{\# total params.}&790,384&1,140,041&1,488,881\\
    \bottomrule
  \end{tabular}
\end{table}

\begin{table}
  \caption{The JSD is upper bounded by $\log S$, and the higher it is the more diverse the mixtures/ensembles are. As expected, the mixtures are the most diverse.}
  \label{tab: jsd}
  \vspace{-2.7 mm}
  \centering
  \begin{tabular}{lllll}
    \toprule

    Model  & $S=2$ & $S=3$     & $S=4$ & $S=5$  \\
    \midrule
    
    Ensemble VAE  &  $0.31$ & $0.46$ & $0.54	$ & $0.59$ \\
    Mixture VAE  & $\mathbf{0.45}$ & $\mathbf{0.70}$ & $\mathbf{0.89}$ & $\mathbf{1.01}$\\
    
    $\log S$  &  $0.69$ & $1.10$ & $1.39	$ & $1.61$ \\
    \bottomrule
  \end{tabular}
\end{table}

\begin{table*}
  \caption{\textbf{scVI:} Accuracy for linear classification using the latent representations for an increasing number of mixture components ($S$).}
  \label{tab: scvi 2}
  \centering
  \begin{tabular}{llllll}
    \toprule

    Model     & $S=1$     & $S=2$ & $S=3$     & $S=4$ & $S=5$  \\
    \midrule
    VAE (scVI)  &  94.2$\pm$.02\% & 96.1$\pm$.02\% & 95.9$\pm$.01\% & 95.5$\pm$.01\% & 96.1$\pm$.02\%\\ 
    
    Ensemble VAE  &  \textbf{94.7$\pm$.02}\% & \textbf{96.4$\pm$.02}\% & \textbf{96.2$\pm$.01}\% & 96.5$\pm$.02\% & 96.8$\pm$.01\%\\
     
    Mixture VAE  &  \textbf{94.7$\pm$.02}\% & 95.9$\pm$.01\% & \textbf{96.2$\pm$.01}\% & \textbf{97.0$\pm$.01}\% & \textbf{97.1$\pm$.01}\%\\
    \bottomrule
  \end{tabular}
\end{table*}

\subsection{Latent Representations}
Here we evaluate the quality of the latent representations produced by Mixture VAEs and compare it to the Vanilla VAEs and Ensemble VAEs. Our evaluation is quantitative and, as is common \citep{sinha2021consistency,locatello2019challenging}, assessed by performing a linear classification on the representations. We use two diverse types of data, single cell data and images. The linear classifier was a linear SVM from the Scikit library \citep{pedregosa2011scikit}.

A VAE provides the practitioner with mainly two alternatives for representing $x$ in lower dimensions, either via sampling $z$ conditioned on $x$ \citep{lopez2018deep} or via the parameters of the approximate distribution \citep{dieng2019avoiding}. Encouraged by our results in Sec. \ref{sec:2d experiments}, we hypothesize that the Mixture VAE learns complex representations of the data by coordinating its components such that each component contributes in a non-trivial way. To exploit this, we feed the mixture parameters into the classification nets. For the baselines that use single Gaussian approximations, we instead sample multiple times from $q_\phi(z|x)$, such that the number of training features for the classifiers is the same.

\subsubsection{Images}
\label{sec:latent images}
The VAE architectures here are two-layered MLP versions from \citep{tomczak2018vae}, and are trained according to the same scheme as described in Sec. \ref{sec:app ll}. The experiment is carried out on the MNIST and FashionMNIST datasets. In addition to the obtained accuracies, we also quantify the diversity of the mixtures and ensembles in terms of the general Jensen-Shannon divergence (JSD) for uniform mixtures, using Eq. \eqref{eq:jsd}. 
The JSD is non-negative, upper bounded by $\log S$, and is maximized when all components have non-overlapping supports.

From Tables \ref{tab: mnist} and \ref{tab: fashionmnist}
it is clear that the Mixture VAEs are superior on this task for all $S>1$, and in Table \ref{tab: jsd} we show that the Mixture VAEs obtains more diverse approximations than the ensemble VAE as measured by in JSD on the MNIST dataset. Using the JSD results, we are also able to explain the stagnating accuracies by the Ensemble VAEs. The JSDs are barely increasing with $S$, meaning that more components are not necessarily contributing with more information. 


\subsubsection{Single-cell data}
The scVI algorithm \citep{lopez2018deep} is a deep generative model for single-cell data, and is a state-of-the-art VAE algorithm that incorporates a probabilistic model for single-cell transcriptomics. We implement it using scVI-tools \citep{Gayoso2022}. This experiment is carried out on the mouse cortex cells dataset (CORTEX) \citep{zeisel2015cell}, which consists of 3005 cells and labels for seven distinct cell types. We used the top 1.2k genes, ordered by variance. Table \ref{tab: scvi 2} shows the accuracy results for the linear classifier on the latent representations from scVI. Mixture VAE outperforms the baselines for $S>3$. However, the Ensemble VAE is also able to utilize its representations for lower $S$. where it performs the best. In Sec. \ref{sec:app sc}, we show the supremacy of Mixture VAEs on clustering tasks.

 \section{Related Work}
 Mixture VAEs have previously appeared in the literature as a means to obtain more flexible posterior approximations \citep{pires2020variational, morningstar2021automatic}. As far as we are aware, they appeared first in \citet{nalisnick2016approximate}, who introduced a Gaussian mixture model latent space and placed a Dirichlet prior on the mixture weights. The mixture weights for the mixture approximation were then inferred via a neural network mapping from $x$ to the simplex. In another example \citep{roeder2017sticking}, a weighted mixture ELBO, more similar to MISELBO with $L=1$, was used to train a VAE. They showed that stop gradients could be applied to Mixture VAEs, and discussed how to sample from the mixture in order to approximate the ELBO. The latter Mixture VAE was only applied to a toy data set. Although these are important works on Mixture VAEs, there are no earlier examples in the literature where the general applicability of Mixture VAEs has been demonstrated. We show that Mixture VAEs can be off-the-shelf tools, and provide the \textit{mixture cookbook}, the first of its kind.

 In the multimodal VAE literature \citep{shi2019variational, sutter2021generalized, javaloy2022mitigating}, mixtures of variational approximations have been studied in depth. A clear distinction to the mixture VAEs considered here is that, in the multimodal VAE works, there are $M$ ``experts'', where $M$ is determined by the number of modalities in the data. Each expert is a variational mixture component conditioned on a certain data modality. The insights in this work can, however, be leveraged in multimodal VAEs by modelling each expert as a mixture distribution.
 
 Outside the VAE literature, mixture learning with variational objectives have been presented in, for example, automatic differentiation VI \citep{kucukelbir2017automatic} and Thompson sampling based bandits \citep{wang2020thompson}. Finally, our composite model builds on the architectures presented in \citet{tomczak2018vae} and \citep{kingma2016improved}, however, our model is a novel composition of the two methods applied to Mixture VAEs.

\section{Conclusions}
In this work, we have shown that mixture components cooperate to cover the target distribution when learned with MISELBO, and connect this to the AIS methodology. Furthermore, we have made Mixture VAEs more accessible by providing the mixture cookbook, which shows how to train mixtures of encoders for advanced VAE methods. Finally, we put forth a novel Mixture VAE, the composite model, which outperforms most of the state-of-the-art methods on the MNIST dataset.

\section{Future Work}
Our work promotes future research on Mixture VAEs. For instance, do the mixture approximations result in better generative models (decoders) in VAEs? Since the decoder is learned by maximizing MISELBO with respect to $\theta$, we expect the quality of the decoder to be affected by the other factors in MISELBO, that is, the variational distribution. This needs further investigation, potentially using the insight in \citet{wu2016quantitative}. In this work, the monotonically non-increasing NLL values were most apparent when the Pixel-CNN decoder was used, i.e. in the MNIST and FashionMNIST experiments.

In this work, separate encoders were used to model the mapping from $x$ to the parameters of the mixture components, making the number of network parameters increase in a way that may not be acceptable in certain applications. A sensible research direction is thus to investigate whether shared encoder network weights can produce similar results as those demonstrated here.

Another interesting path is to conduct more real data experiments, investigating the usefulness of mixtures in VI in practice. Evolutionary biology has recently seen a trend in employing VI for phylogenetic tree inference \citep{zhang2018variational, koptagel2022vaiphy}. Given the results displayed here, mixtures could help VI algorithms to explore the large and complex phylogenetic tree spaces, which is a known difficulty in Bayesian phylogenetics.

\section*{Acknowledgments}
First, we acknowledge the insightful comments provided by the reviewers, which have helped
improve our work. This project was made possible through funding from the Swedish Foundation for
Strategic Research grants BD15-0043 and ID19-0052, and from the Swedish Research Council grant 2018-05417\_VR.
The computations and data handling were enabled by resources provided by the Swedish National
Infrastructure for Computing (SNIC), partially funded by the Swedish Research Council through
grant agreement no. 2018-05973.

\bibliography{example_paper}
\bibliographystyle{icml2023}

\newpage
\appendix
\onecolumn

\section{Two Dimensional Experiments}
\label{sec:app 2d experiments}
In both experiments, we consider a complicated $p(z)$. All approximations were initialized at $z_1 = z_2 = 0$ and their parameters were optimized using the ADAM optimizer \citep{kingma2014adam} with a learning rate $0.1$. 

The objective function to be minimized was either the KL divergence between the approximation and $p(z)$, or, for the single approximation with $L=30$, an importance-weighted variant of the KL 
\begin{equation}
    \mathcal{D}_L = -\mathbb{E}_{z_\ell \sim q_{\phi}(z)}\left[
    \log\frac{1}{L}\sum_{\ell=1}^L\frac{p(z_{\ell})}{ q_{\phi}(z_{\ell})}
    \right].
\end{equation}
We believe that the effects of minimizing $\mathcal{D}_L$ are not well studied, although, one instance of $\mathcal{D}_L=0$ is clearly when the approximation exactly matches $p(z)$. Future work will hopefully bring interesting insights and enable us to explain the behavior exhibited in the experiments by the importance weighted approximation (IWVI).

All approximations were initialized at $(0, 0)$, but the same behavior/solutions were obtained for the mixture approximation for alternative starting points. In some cases, a reduced learning rate helped the mixture approximations.

Finally, when plotting $p(z)$ we uniformly sampled $z$ in a square that captured their supports. We then plotted the points with a likelihood of more than $0.1$, i.e. $p(z)>0.1$.

\section{Log-Likelihood Estimation Details}
\label{sec:app ll}
For MNIST \citep{lecun-mnisthandwrittendigit-2010} and FashionMNIST \citep{xiao2017fashion} we used $p_\theta(x|z) = \text{Bernoulli}(x; \theta)$, assuming independence across the pixels, while a 3-component mixture of discretized Logistic distributions \citep{salimans2017pixelcnn} was used for CIFAR-10 \citep{krizhevsky2009learning}. To implement the mixture of discretized Logistic distributions in PyTorch, we used the provided code from \citet{vahdat2020nvae}. Otherwise, our code was in general largely inspired by the one provided by \citet{tomczak2018vae}.

As is standard, we computed the bits per dimension scores as
\begin{equation}
    \textnormal{BPD}(x; \theta) = \frac{\log p_\theta(x)}{d_x \log(2)},
\end{equation}
where $d_x=32^2\cdot 3$ for CIFAR-10 and the marginal log-likelihood was approximated via MISELBO.

Regarding data augmentation, we used horizontal flips for the CIFAR-10 training data (as in \citet{vahdat2020nvae}), and Bernoulli draws for the pixel values for the MNIST and FashionMNIST data (as in \citet{tomczak2018vae}).

To approximate the marginal log-likelihood, we used $L=5000$ and $L=100$ importance samples for MNIST/FashionMNIST and CIFAR-10, respectively. These choices were based on examples in the literature \citep{tomczak2018vae, vahdat2020nvae}.

For all models when training on the MNIST or FashionMNIST datasets, we utilized a linearly increasing KL warm-up \citep{sonderby2016ladder} for the first 100 epochs, early stopping with 100 epochs look ahead, and the ADAM optimizer \citep{kingma2014adam} with a learning rate of $0.5\cdot10^{-4}$ and no weight decay. Observe that this is the training scheme used in \citet{tomczak2018vae} but with a longer look ahead. For CIFAR-10 we employed an identical scheme but trained for a maximum of 1000 epochs with no look-ahead mechanism. In all setups here, we used a training and validation batch size of 100.

The models used for MNIST and FashionMNIST had 40-dimensional latent spaces, while the CIFAR-10 related models had 124-dimensional latent spaces (as in \citet{thin2021monte}). Furthermore, the ``vanilla'' VAEs used on CIFAR-10 had CNN encoder networks along with a PixelCNN decoder.

For all large-scale experiments, we used a single 32-GB Tesla V100 GPU, or a desktop computer with a 10-GB RTX 3080 GPU. 

As mentioned in Sec. \ref{sec:components}, during the training of the models we employed KL warm-up. For the vanilla Mixture VAE, this results in the following objective function
\begin{equation}
    \mathcal{L}^\beta_\textnormal{MIS} = \frac{1}{S}\sum_{s=1}^S \mathbb{E}_{q_{\phi_s}}\left[
    \log  p_\theta(x| z_{s})\right] + \beta\frac{1}{S}\sum_{s=1}^S  \mathbb{E}_{q_{\phi_s}}\left[ \log\frac{p_\theta(z_s)}{\frac{1}{S}\sum_{j=1}^S q_{\phi_j}(z_{s}|x)}
    \right],
\end{equation} 
where $\beta$ linearly increased from zero to one over the first 100 epochs. 

Unlike NVAE \citep{vahdat2020nvae}, to name another large VAE model, our models were trained on single GPU nodes. Parallelizing the encoder operations, for instance, over multiple GPUs is easily achieved and will speed up the training process considerably. The $S=1$ composite model trained for 9 hours, while the biggest model, the $S=4$ composite model, trained for 52 hours and had 5.2 million network parameters. For more info regarding the NVAE architecture, hyperparameters and run time, see here \url{https://github.com/NVlabs/NVAE#running-the-main-nvae-training-and-evaluation-scripts}..

Finally, we provide the formulation of MISELBO with $L=1$ for the composite model
\begin{equation}
\label{eq:miselbo_composite}
    \mathcal{L}_\textnormal{MIS} = \frac{1}{S} \sum_{s=1}^S \mathbb{E}_{z_{s}^{1},z_{s}^2 \sim q_{\phi_s}(\mathbf{z_s}|x)}\left[
    \log \frac{p_\theta(x| z^{T,1}_{s}, z^2_{s})p_\theta(z^{T,1}_{s}|z^2_s)\frac{1}{KS} \sum_{k=1}^K\sum_{j=1}^S q_{\phi_j}(z^2_s|u_{k})
    }{\frac{1}{S}\prod_{t=1}^T\left\vert \frac{df_{\gamma^t}(z_{s}^{t-1, 1})}{dz_{s}^{t-1, 1}}\right\vert^{-1}\sum_{j=1}^S q_{\phi^2_j}(z^2_s|x)q_{\phi^1_j}(z^{0,1}_{s}|z^2_s,x)}
    \right].
\end{equation}
In this work we used uniform mixture weights, however the expression above can trivially be extended to non-uniform weights.

\begin{figure}
    \centering
\includegraphics[width=0.8\textwidth]{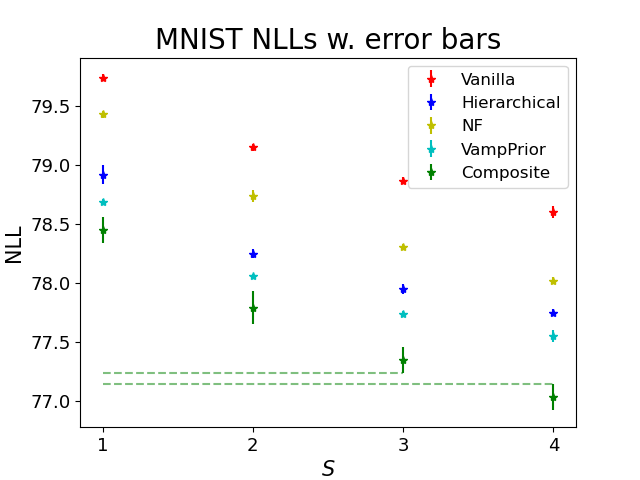}
    \caption{Error bars using a single standard deviation for the NLL scores on the MNIST dataset. Note that none of the error bars are overlapping for any model as $S$ increases. This is emphasized with the dashed green line, showing that the composite model's results for $S=3$ and $S=4$ do not overlap.}
    \label{fig:mnist errorbars}
\end{figure}

\section{Big IWAE versus Mixture VAE}
\label{sec:big_mix_appendix}
Indeed, the number of encoder network parameters increases as a function of $S$, and the Mixture VAE uses $S\times L$ samples to approximate its objective function. Hence, it is necessary to explore whether the improved NLL scores by the Mixture VAEs are due to the mixture approximations, or due to the increased number of parameters and the number of importance samples. In general, we found that the Big IWAE could not outperform the corresponding Mixture VAE regardless if we increased the number of hidden layers or the number of hidden units. 

In order to restrict the hyper-parameter space, we perform this analysis in isolation from the experiments in  the main text and consider simpler VAEs. Namely, we employ the vanilla MLP VAE from \citep{tomczak2018vae}, without the gating mechanisms. This gives the encoder networks of the Mixture VAE two hidden layers with 300 hidden units each. We trained the Mixture VAEs with $S=1,2,3$ and $L=1$, and we adjusted the number of parameters in the encoder networks of the importance weighted autoencoder (IWAE; \citet{burda2015importance}) to get a corresponding number of parameters with $L=S$.  

In general, we found that the Big IWAE could not outperform the corresponding Mixture VAE regardless if we increased the number of hidden layers or the number of hidden units. In most cases, we found the opposite; more parameters in the Big IWAE resulted in worse NLL. This does not mean that single encoder networks cannot be scaled to outperform Mixture VAEs in general, but it shows that naively increasing the number of parameters does not necessarily lead to better log-likelihood estimates. On the other hand, naively adding mixture components work well in this regard. 

To make the negative log-likelihood (NLL) comparison between Big IWAE and the mixture VAE fair, we estimated the marginal log-likelihood by performing Monte Carlo sampling either via
\begin{equation}
    \log p_\theta(x) \approx \frac{1}{S}\sum_{s=1}^S \log \frac{1}{L}\sum_{\ell=1}^L \frac{p_\theta(x, z_{s,\ell})}{\frac{1}{S}\sum_{j=1}^Sq_{\phi_j}(z_{s,\ell}|x)}, \quad z_{s,\ell} \sim q_{\phi_s}(z|x),
\end{equation}
if the variational approximation is a mixture distribution, or 
\begin{equation}
\label{eq:is approx}
    \log p_\theta(x) \approx \frac{1}{S}\sum_{s=1}^S \log \frac{1}{L}\sum_{\ell=1}^L \frac{p_\theta(x, z_{s,\ell})}{q_{\phi}(z_{s,\ell}|x)}, \quad z_{s,\ell} \sim q_{\phi}(z|x),
\end{equation}
in the $S=1$ case. That is, MISELBO and IWELBO were both approximated by sampling $L$ importance samples $S$ times. We report the scores in terms of NLL which were obtained using $L=5000$ following \citet{tomczak2018vae}.

\section{Latent Representations}
\subsection{Image data}
The expression for the JSD is given here,
\begin{equation}
\label{eq:jsd}
    \text{JSD}(\{q_{\phi_s}\}_{s=1}^S) = \mathbb{H}\left[
    \frac{1}{S}\sum_{s=1}^S q_{\phi_s}(z|x)
    \right] - \frac{1}{S}\sum_{s=1}^S\mathbb{H}\left[
    q_{\phi_s}(z|x)
    \right].
\end{equation}

\begin{table}
  \caption{\textbf{FashionMNIST:} Accuracy for linear classification using the latent representations for an increasing number of mixture components ($S$).}
  \label{tab: fashionmnist}
  \centering
  \begin{tabular}{llllll}
    \toprule
    Model     & $S=1$     & $S=2$ & $S=3$     & $S=4$ & $S=5$  \\
    \midrule
    
    VAE  &  79.5$\pm$3E-3\% & 79.7$\pm$3E-5\% & 79.6$\pm$2E-3\% & 79.7$\pm$2E-3\% & 79.5$\pm$4E-3\%\\
    

    Ensemble VAE   & \textbf{80.6$\pm$2E-3}\% &  81.1$\pm$3E-3\% & 81.7$\pm$1E-3\% & 82.5$\pm$6E-4\% & 82.6$\pm$1E-3\% \\

    Mixture VAE & \textbf{80.6$\pm$2E-3}\% & \textbf{81.8$\pm$2E-3}\% & \textbf{82.3$\pm$1E-3}\% & \textbf{82.7$\pm$1E-3}\% & \textbf{83.5$\pm$2E-3}\%\\

    
    \bottomrule
  \end{tabular}
\end{table}

\subsection{Single-Cell Data}
\label{sec:app sc}
We have implemented and trained scVI models based on scVI-tools \citep{Gayoso2022}. We have randomly split CORTEX data into train, validation, and test sets with the following ratio: 80\%, 7\%, 13\%. Each model has been trained 1000 epochs with an early stop condition of having non-decreasing loss for 100 epochs. All the classification/clustering experiments were performed with the latent representations of the test set.

In the scVI latent space, we also do clustering using $k$-means from the Scikit-learn library \citep{pedregosa2011scikit}. Then following \citep{lopez2018deep}, we quantitatively assess the quality of clustering by applying two clustering metrics: adjusted rand index (ARI) and normalized mutual information (NMI). The results below show that the latent representations produced by the Mixture VAE extension of scVI perform better compared to the latent representations produced by the vanilla and Ensemble VAE extensions of scVI.

We assess the quality of latent features using two clustering metrics, which are employed by scVI \citep{lopez2018deep} to compare against baselines. These metrics are adjusted rand index (ARI) and normalized mutual information (NMI). 
\par \textit{Adjusted rand index (ARI):} is the rand index adjusted for chance and used as a similarity measure between the predicted and true clusterings. It is mathematically defined as:
\begin{equation*}
    \label{eqn: ARI}
    ARI = \frac{\sum_{ij} \binom{n_{ij}}{2} - \Big[\sum_i \binom{a_i}{2} \sum_j \binom{b_j}{2} \Big] \Big/ \binom{n}{2}  } {\frac{1}{2} \Big[\sum_i \binom{a_i}{2} + \sum_j \binom{b_j}{2} \Big] - \Big[\sum_i \binom{a_i}{2} \sum_j \binom{b_j}{2} \Big] \Big/ \binom{n}{2}   }
\end{equation*}
where $n_{ij}$, $a_i$, $b_j$ are values from the contingency table. The higher ARI score means the more the two clusterings agree. ARI can be at most 1, which indicates that the two clusterings are identical.

\begin{table}[h]
  \caption{ARI: adjusted rand index (higher is better)}
  \label{tab: scvi clustering}
  \centering
  \begin{tabular}{llllll}
    \toprule

    Model     & $S=1$     & $S=2$ & $S=3$     & $S=4$ & $S=5$  \\
    \midrule
    VAE (scVI)  &  $0.82$ & $0.82$ & $0.82$ & $0.81$ & $0.82$\\ 
    
    Ensemble VAE  &  $0.82$ & $0.83$ & $0.83$ & $0.82$ & $0.83$\\
     
    Mixture VAE  &  $0.82$ & $0.82$ & $0.84$ & $0.85$ & $0.85$\\
    \bottomrule
  \end{tabular}
\end{table}

\textit{Normalized Mutual Information (NMI):} is the normalized version of MI score, which is a similarity measure between two different labels of the same data. Normalization is done in a way so that the results would be between 0 (no mutual information) and 1 (perfect correlation): 
\begin{equation*}
    \label{eqn: NMI}
    NMI = \frac{I(P;T)}{\sqrt{\mathbb{H}(P)\mathbb{H}(T)}}
\end{equation*}
where $P$ and $T$ denote the empirical categorical distributions of the predicted and true labels, respectively. $I$ is the mutual entropy, and $\mathbb{H}$ is the Shannon entropy. The higher NMI score indicates a better match between the two label distributions. 

\begin{table}[h]
  \caption{NMI: normalized mutual information (higher is better)}
  \label{tab: scvi clustering NMI}
  \centering
  \begin{tabular}{llllll}
    \toprule

    Model     & $S=1$     & $S=2$ & $S=3$     & $S=4$ & $S=5$  \\
    \midrule
    VAE (scVI)  &  $0.81$ & $0.81$ & $0.81$ & $0.81$ & $0.81$\\ 
    
    Ensemble VAE  &  $0.82$ & $0.82$ & $0.82$ & $0.81$ & $0.81$\\
     
    Mixture VAE  &  $0.82$ & $0.82$ & $0.83$ & $0.84$ & $0.84$\\
    \bottomrule
  \end{tabular}
\end{table}

\end{document}